\def\year{2019}\relax
\documentclass[letterpaper]{article} 
\usepackage{aaai19}  
\usepackage{times}  
\usepackage{helvet}  
\usepackage{courier}  
\usepackage{url}  
\usepackage{graphicx}  

\usepackage{amsmath} 
\usepackage{multirow}
\usepackage{pinyin}
\usepackage{CJK} 
\usepackage{color}
\usepackage{booktabs}
\usepackage[normalem]{ulem}

\begin{document}
%
\title{Language-Independent Representor for Neural Machine Translation}
\author{Long Zhou$^{1,2}$, Yuchen Liu$^{1,2}$, Jiajun Zhang$^{1,2}$, Chengqing Zong$^{1,2,3}$, Guoping Huang$^4$\\
	$^1$National Laboratory of Pattern Recognition, CASIA, Beijing, China \\
	$^2$University of Chinese Academy of Sciences, Beijing, China \\
	$^3$CAS Center for Excellence in Brain Science and Intelligence Technology, Shanghai, China \\
	$^4$Tencent AI Lab \\
	\{long.zhou, yuchen.liu, jjzhang, cqzong\}@nlpr.ia.ac.cn, 
	donkeyhuang@tencent.com \\
}
\maketitle

\begin{abstract}
  Current Neural Machine Translation (NMT) employs a language-specific encoder to represent the source sentence and adopts a language-specific decoder to generate target translation. This language-dependent design leads to large-scale network parameters and makes the duality of the parallel data underutilized.
  To address the problem, we propose in this paper a language-independent representor to replace the encoder and decoder by using weight sharing. This shared representor can not only reduce large portion of network parameters, but also facilitate us to fully explore the language duality by jointly training \textbf{source-to-target}, \textbf{target-to-source}, \textbf{left-to-right} and \textbf{right-to-left} translations within a multi-task learning framework.
  Experiments show that our proposed framework can obtain significant improvements over conventional NMT models on resource-rich and low-resource translation tasks with only a quarter of parameters.
\end{abstract}

\section{Introduction}

End-to-end neural machine translation (NMT) has significantly improved the quality of machine translation in recent several years~\cite{Bahdanau:2015,gehring2017convolutional,vaswani2017attention}.
Although NMT has shown superior performance on public benchmarks~\cite{bojar2017findings} and rapid adoption in deployments by, e.g., Baidu~\cite{zhou2016deep}, Google~\cite{Wu:2016}, and Microsoft~\cite{hassan2018achieving},
it still faces many challenges~\cite{koehn2017six}.

No matter which basic blocks we use, such as recurrence~\cite{Bahdanau:2015}, convolution~\cite{gehring2017convolutional}, or self-attention~\cite{vaswani2017attention},
conventional NMT adopts language-specific encoder to transform the source language and utilizes a language-specific decoder to generate target translation token by token.
It is clear that this language-dependent encoder-decoder framework has two problems. First, encoder and decoder have similar structures but contain separate parameters, resulting in the waste of enormous parameters. Second, one NMT model can only perform one unidirectional translation task using parallel corpora, which cannot take full advantage of language duality.

To address the issue of parameters scale, existing studies usually use weight pruning or knowledge distillation~\cite{see2016compression,kim2016sequence} to compress NMT models.
\citeauthor{press2016using}~\shortcite{press2016using} conducted the weight tying of input and output embedding in RNN-based NMT.
However, their researches are built under the language-dependent encoder-decoder framework which does not consider the language commonality. 
Being orthogonal to previous work, we are interested in exploiting language-independent model to reduce network parameters.

On the other hand, dual properties and agreement of translation have attracted much attention in  NMT~\cite{cheng2015agreement,liu2016agreementa,tu2017neural,hassan2018achieving}.
\citeauthor{cheng2015agreement}~\shortcite{cheng2015agreement} proposed  agreement-based joint training for source-to-target and target-to-source translation directions. Besides, \citeauthor{liu2016agreementa}~\shortcite{liu2016agreementa} focused on the agreement between left-to-right and right-to-left on the target side to overcome the unbalanced output problem.
Although these approaches have incorporated language duality into NMT, they all use language-specific encoders and decoders for each language and each direction separately.
How to integrate multiple translation tasks into one model is still an open question.

\begin{figure}
\setlength{\belowcaptionskip}{-0.3cm}
    \centering
    \includegraphics[width=8cm]{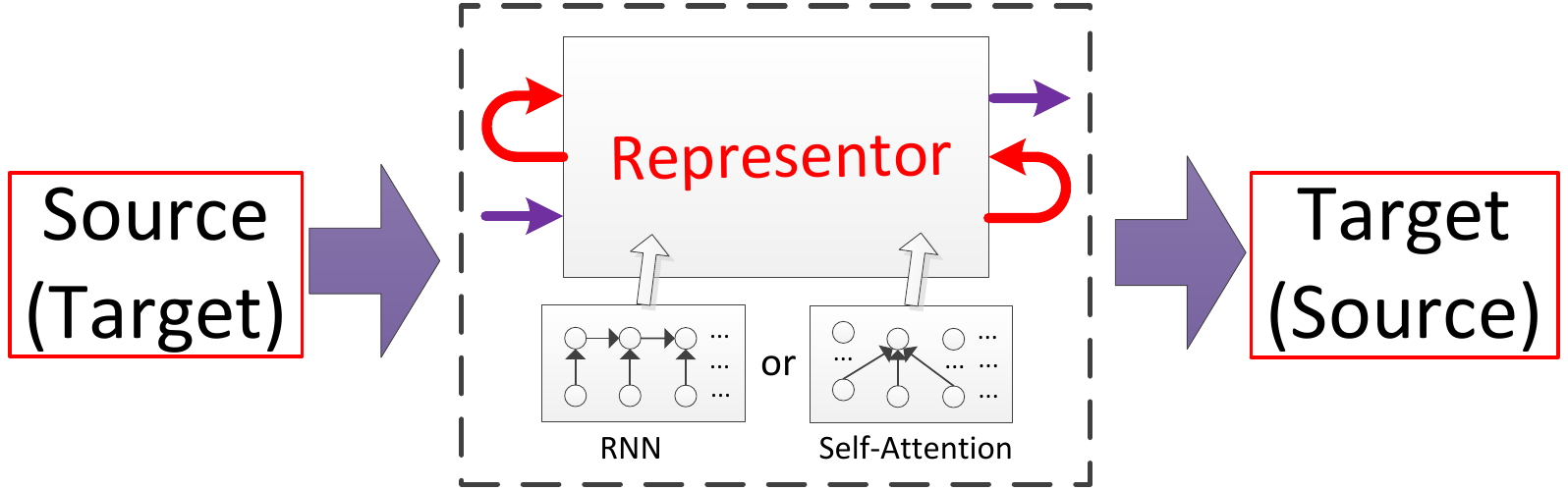}
    \caption{Our language-independent representor, replacing encoder and decoder in conventional NMT, can perform and combine the advantage of source-to-target and target-to-source translation tasks by leveraging language duality.
    }\label{representor}
\end{figure}


In this work, we introduce a simple yet highly effective language-independent NMT framework by using weight sharing and multi-task learning.
To reduce model parameters, we first present a language-independent representor by investigating the effectiveness of weight sharing in different hierarchies, including embeddings weight sharing, layer weight sharing and encoder-decoder sharing.
With the ability of representing both source and target languages, the shared representor inspires us to make full use of the language duality by jointly training  source-to-target, target-to-source, left-to-right, and right-to-left translations within a multi-task learning framework.
We verify the effectiveness of this framework on resource-rich Chinese$\leftrightarrow$English and low-resource English$\leftrightarrow$Japanese translation tasks.
Experimental results demonstrate that our model can leverage only a quarter of parameters to achieve substantial improvements over conventional NMT models.

Specifically, we make the following contributions in this paper:
\begin{itemize}
\item To the best of our knowledge, this is the first work to introduce a language-independent representor to replace encoder and decoder  in conventional NMT by using weight sharing. Specifically, this framework achieves model compression from a very different perspective.
\item Our model can perform and combine the advantage of four translation tasks in a representor by utilizing language duality, which contains \emph{source-to-target}, \emph{target-to-source}, \emph{left-to-right}, and \emph{right-to-left} translations.
\item Our proposed framework drastically reduces model parameters and achieves significant improvements especially for low-resource translations, where our framework can be viewed as a data augmentation technique.
\end{itemize}

\section{Background}

Both RNN-based NMT (RNMT)~\cite{Luong:2015A} and Transformer~\cite{vaswani2017attention} employ a language-dependent encoder-decoder structure, consisting of stacked encoder and decoder layers.
The encoder maps an input sequence of symbol representations $(x_1, ..., x_m)$ to a sequence of continuous representations $z=(z_1, ..., z_m)$. Given $z$, the decoder then generates an output sequence $(y_1, ..., y_t)$ of symbols one element at a time.
At each step the model is auto-regressive, consuming the previously generated symbols as additional input when generating the next.

\begin{figure}
\setlength{\belowcaptionskip}{-0.3cm}
    \centering
    \includegraphics[height=9cm]{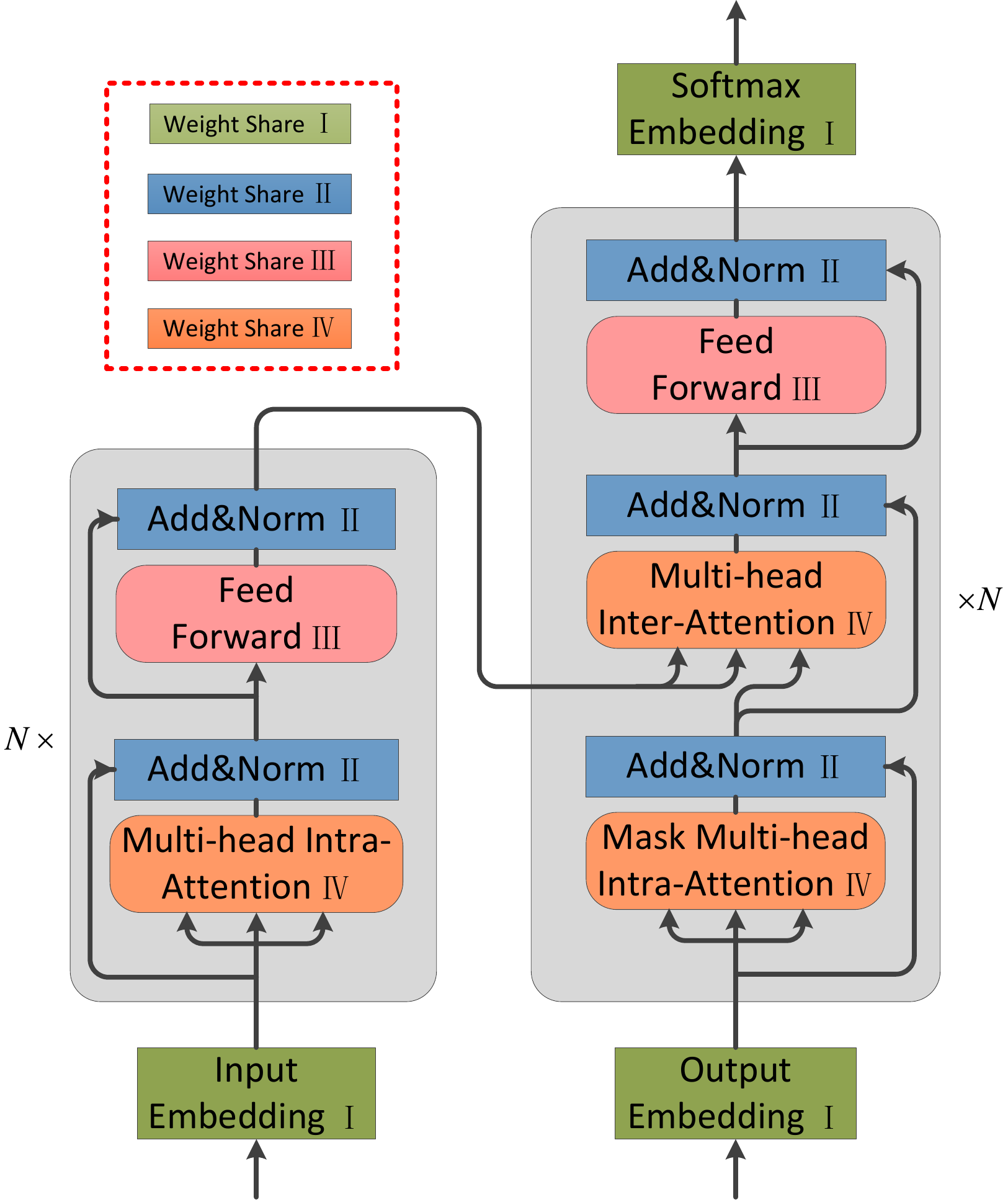}
    \caption{The architecture of Transformer, also of our representor, which only has four components for Transformer architecture: embedding, multi-head attention, feed-forward and layer normalization weight matrices.
     The same Roman number or color indicates weight sharing among them.
    } \label{transformer-share}
\end{figure}

Here, we mainly introduce the Transformer, as shown in Figure \ref{transformer-share}.
Encoder layers consist of two sublayers: multi-head intra-attention followed by a position-wise feed-forward layer. Decoder layers consist of three sublayers: multi-head intra-attention followed multi-head inter-attention, and then followed by a position-wise feed-forward layer.
It uses residual connections around each of the sublayers, followed by layer normalization.
The decoder uses masking in its self-attention to preserve the auto-regressive property during training step.


Given a set of training examples $\{x^{(n)}, y^{(n)}\}^N_{n=1}$, the training algorithm aims to find the model parameters that maximize the likelihood of the training data:
\begin{equation}
    J(\theta) = \sum_{n=1}^N log P(y^{(n)}|x^{(n)}; \theta)
\end{equation}
For the sake of brevity, we refer the reader to \cite{Luong:2015A} and \cite{vaswani2017attention} for additional details regarding the architecture.

\section{Our Approach}

\begin{figure*}
\setlength{\belowcaptionskip}{-0.5cm}
    \centering
    \includegraphics[height=8.6cm]{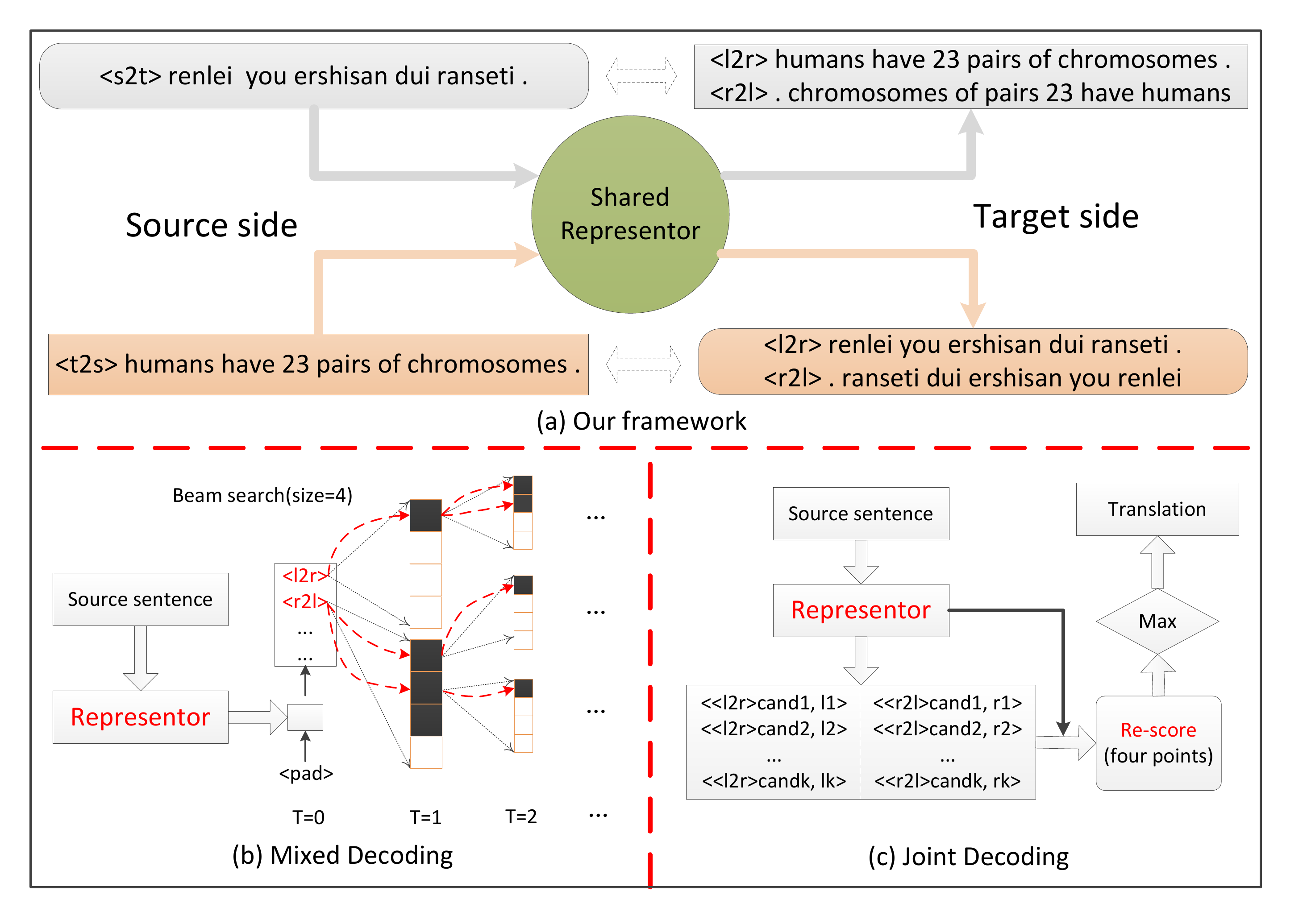}
    \caption{Our proposed framework and two decoding methods.
    Our goal is to perform multiple translation tasks in a language-independent representor by utilizing language duality, as shown in (a) for Chinese-English translation.
    (b) denotes the mixed decoding which can combine the left-to-right decoding and right-to-left decoding in one beam-search process determined by model.
    Joint decoding (c) works as a reranking technique to select a better translation from the two k-best candidates generated by the decoder.
    }\label{fig:1}
\end{figure*}

We introduce a simple language-independent NMT framework which can not only reduce the model parameters, but also take full advantage of language duality.
The central idea is to achieve language-independent representor by weight sharing, and perform multiple translation tasks by multi-task learning, as shown in Figure~\ref{fig:1}(a).

\subsection{Language-Independent Representor} \label{weightshare}

Transformer and RNMT still employ language-dependent encode-decoder framework which leads to large-scale parameters. In this section, we will introduce a language-independent representor achieved by embedding weight sharing and encoder-decoder sharing, as shown in Figure \ref{transformer-share}.

{\bf Embedding Weight Sharing (ES):} NMT model uses embeddings to convert input tokens and output tokens to vectors.
It also utilizes the linear transformation and softmax function to convert the decoder output to next-token probabilities.
\citeauthor{press2016using}~\shortcite{press2016using} conducted the weight tying of input and output embedding in RNMT, whose results show that it can reduce the size of NMT models without harming their performance.
However, they conducted the experiments on English-German and English-French parallel pairs which are similar languages and have shared source-target vocabulary. What if they are not similar languages, such as Chinese-English or Japanese-English translation?

To address the problem, we design a \emph{\textbf{frequency-based embedding weight sharing}} strategy. The steps are as follows: (1) We first count the frequency of word occurrences in bilingual languages and sort them in descending order.
(2) We select a vocabulary according to the predefined vocabulary size. For example, we limit the source and target vocabularies to the most frequent 35K and 30K tokens, respectively.
(3) Words of bilingual vocabulary having the same frequency rank share a word embedding. For instance, the sixth word of the source vocabulary and the sixth word of the target vocabulary use the same word embedding in training and inference.
(4) If the source and target vocabularies have different vocabulary sizes, the length of shared embedding weight matrix is maximum of the source and target vocabulary size.
As a result, the input embedding of the decoder, the output embedding of the decoder and the input embedding of the encoder are all shared.

{\bf Encoder-Decoder Sharing (EDS):} Encoder and decoder play a key role in NMT, which map sequence of symbol representation to sequence of continuous representation.
Weight sharing, such as RNN weight matrices or multi-head attention weight matrices, has been used for the different time steps within encoder or decoder.
Here, we mainly focus on the weight sharing between encoder and decoder.

As shown in Figure~\ref{transformer-share}, Transformer has some similar components which play analogous roles in every layer, consisting of multi-head attention layer, feed-forward layer, and layer normalization layer.
The same Roman number or color indicates weight sharing among them in every layer.
In other words, our representor replaces encoder and decoder of conventional NMT, and it only has three sets of parameters, namely self-attention, feed-forward and layer normalization weight matrix.
Additionally, we also explore the \textbf{Layer Weight Sharing (LS)}, in which every layer of encoder and decoder shares common weights .
For RNMT, our proposed representor adopts one RNN (LSTM~\cite{hochreiter1997long} or GRU~\cite{Cho:2014}) to replace encoder and decoder.

\subsection{Multi-Task Learning for Representor} \label{multi-tast}

In the last section, we present a language-independent representor which can reduce large portion of network parameters.
With the ability of representing both source and target languages, the shared representor facilitates us to fully explore the language duality consisting of source-to-target, target-to-source, left-to-right and right-to-left translation.
In this section, we attempt to conduct multiple translation tasks in the shared representor by leveraging multi-task learning framework.

{\bf Source-to-Target and Target-to-Source (S-T\&T-S): }Conventional NMT directly models the probability of a target-language sentence given a source-language sentence. Some work have noticed the symmetry of translation ~\cite{Cheng:2016,he2016dual,tu2017neural,sennrich2017university,zhang2018joint}, which bridge source-to-target and target-to-source translation.
However, above approaches use two encoder-decoder models or additional reconstructor, in which encoder and decoder have separate parameters.
Since the encoder and decoder in our model share a set of parameters where both encoder and decoder are capable of mapping source and target languages, a straightforward idea is to utilize the joint training of source-to-target and target-to-source in the language-independent representor.  

We introduce our new training objective as follows:
{\setlength\abovedisplayskip{-0.8pt}
\begin{equation}
\begin{aligned}
\label{t2s-patterns}
  J(\theta) = \sum_{n=1}^N log P(y^{(n)}|x^{(n)}; \theta)
   + \sum_{n=1}^N log P(x^{(n)}|y^{(n)}; \theta)
\end{aligned}
\end{equation}
where $\theta$ is shared model parameters in a single representor.
Note that the objective consists of two parts: source-to-target likelihood and target-to-source likelihood. In this way, our representor is able to translate both source-to-target and target-to-source. 

{\bf Left-to-Right and Right-to-Left (L-R\&R-L): } The decoders of RNN, convolution, or self-attention based usually generate the target words from left to right.
Many studies have pointed out the shortcoming of unidirectional decoding, and proposed some approaches to combine the advantage of left-to-right decoding and right-to-left decoding ~\cite{liu2016agreementa,sennrich2017university,hassan2018achieving}.
Our goal in this work is to find a way to combine the left-to-right and right-to-left decoding in one end-to-end model.
Formally, the training can be written as the following equation:
{\setlength\abovedisplayskip{-0.8pt}

\begin{equation}
\begin{aligned}
\label{r2l-patterns}
  J(\theta) = \sum_{n=1}^N log P(\overrightarrow{y}^{(n)}|x^{(n)}; \theta )
   + \sum_{n=1}^N log P(\overleftarrow{y}^{(n)}|x^{(n)}; \theta)
\end{aligned}
\end{equation}

where $P(\overrightarrow{y}^{(n)}|x^{(n)}; \theta )$ denotes the sequence generation from left to right, and $P(\overleftarrow{y}^{(n)}|x^{(n)}; \theta)$ denotes the generation from right to left.
All models consisting of an encoder, left-to-right decoder, and right-to-left decoder use a shared representor.


{\bf Combining Four Patterns (CFP):} By integrating the above two train objectives, we present a simple yet highly effective multi-task learning approach that utilizes source-to-target, target-to-source, target left-to-right and target right-to-left translation in one representor to enhance the translation agreement. 
To this end, the training algorithm aims to maximize the likelihood of the training data:
\begin{equation}
\begin{aligned}
  J(\theta) = \sum_{n=1}^N log P(\overrightarrow{y}^{(n)}|x^{(n)}; \theta )
   + \sum_{n=1}^N log P(\overleftarrow{y}^{(n)}|x^{(n)}; \theta)  \\
   + \sum_{n=1}^N log P(\overrightarrow{x}^{(n)}|y^{(n)}; \theta)
   + \sum_{n=1}^N log P(\overleftarrow{x}^{(n)}|y^{(n)}; \theta)
\end{aligned}\label{four-patterns}
\end{equation}
where $\theta$ is shared weight for all translation patterns.
Here, our proposed model can conduct four translation tasks for one parallel corpora in a representor, while using half parameters compared to a standard end-to-end model.

\subsection{Training and Testing} \label{train_and_test}

In order to train multiple translation tasks in a represntor, we design a simple yet smart strategy to indicate the predefined translation direction.
More specifically, we utilize two special labels ($\langle s2t \rangle$ and $\langle t2s \rangle$) in the first word of input sentences to guide the translation tasks (source-to-target or target-to-source).
Besides, we employ another two special labels ($\langle l2r \rangle$ and $\langle r2l \rangle$) at the beginning of output sentences to indicate translating from left to right or from right to left.
It is easy to use the stochastic gradient descent algorithm to implement duality-based joint learning since the single translation model in four directions uses the same training data and model parameters.

Once a model is trained, we can use a beam search to conduct mixed decoding simply.
Alternatively, we introduce a joint decoding method for the proposed framework, as shown in Figure~\ref{fig:1}(b) and \ref{fig:1}(c).

\textbf{Mixed Decoding (MD)}  The central idea of mixed decoding is to combine the left-to-right and right-to-left decoding in one beam-search process simultaneously.
More specifically, the first input token of decoder is a $\langle pad \rangle$ whose embedding is all zeros for initialization.
Instead of adding the label at the beginning of output sentences to guide translating from left to right or from right to left, we predict the first output token ($\langle l2r \rangle$ or $\langle r2l \rangle$) determined by the model. And we do not need to do anything until the end-of-sentence flag is predicted.
That is, our model has the ability to choose left-to-right or right-to-left decoding automatically according to the source representation.

\textbf{Joint Decoding (JD)} Inspired by ~\cite{liu2016agreementa} and ~\cite{tu2017neural}, we adapt a joint decoding method to find a translation that approximately maximizes the likelihood score.
For Equation~\ref{four-patterns}, the joint decoding consists of two steps: 1) run beam search for target left-to-right and right-to-left models independently to obtain two k-best lists; 2) rerank the union of two k-best lists using the joint model to get the best candidate.

\begin{table*}
\setlength{\belowcaptionskip}{-0.5cm}
\centering
\begin{tabular}{l|cc||ccccc}
  \hline
   Model                                           &    Params          &  Percent &MT03   &      MT04   &      M05     &      MT06       & AVE       \\
  \hline
  \hline
   RNMT                                              &      157.0M   &  100\%   &40.15     &    40.37      &        37.25     &    37.91        &     38.92     \\
   RNMT + ES     &      97.1M    & 61.8\%  &\textbf{40.75}     &    \textbf{41.20}      &        \textbf{37.98}     &    \textbf{38.99}        &     \textbf{39.73}$\ddagger$     \\
   RNMT + EDS     &      125.0M   &  79.6\%  &38.24     &    38.57      &        36.02     &    35.91        &     37.19     \\
   RNMT + ES + EDS  &      65.0M    &  41.4\%      &37.77     &    37.77      &        35.49     &    35.27        &     36.84     \\
  \hline
  \hline
  Transformer                                            &   270.8M         &    100\%     &  47.27         &     \textbf{48.83}   &      45.71    &  45.14       & 46.74           \\
  Transformer + ES                     &  207.8M      &  76.7\% &  47.63      &     48.32   &      \textbf{47.51}    &    45.31     &    \textbf{47.19}$\ddagger$      \\
  Transformer + EDS
                                                      &  170.0M       &  62.8\%   & 47.13         &       47.02      &         45.96      &  \textbf{45.46}      &   46.39      \\
  \textbf{Transformer + ES + EDS}
                                                      &    107.1M      &  39.5\%   & \textbf{47.97}        &     47.30    &        46.32    &    45.28    &  46.71     \\
  Transformer + ES + EDS + LS
                                                      &  44.1M        &   16.6\%  & 45.04       &       45.51   &          43.46   &    43.70      & 44.43        \\
  Transformer (base)                                  &      60.0M   &  22.1\%  &  45.62     &    45.67      &       44.45     &   44.72       &     45.12     \\
  Transformer (2-2)                                   &      153.2M   &  56.6\%  &44.72     &    46.03      &        43.23     &    42.25        &     44.05     \\
  \hline
\end{tabular}
\caption{Results of our proposed representor for Chinese-English translation.
  Although the number of model parameters is drastically reduced, their performance is comparable to baseline.
  ES and EDS mean embedding weight sharing and encoder-decoder sharing separately.
  ``Transformer + ES + EDS", namely our proposed representor, will be used in latter experiments.
  LS denotes layer weight sharing, and the last line shows the results that transformer has two layers encoder and decoder respectively.
  ``$\ddagger$": significantly better than baseline (p $<$ 0.05).
} \label{resultWS}
\end{table*}

\section{Experimental Settings}

\subsection{Dataset}

We evaluate our experiments on large-scale NIST Chinese$\leftrightarrow$English translation tasks, and low-resource KFTT English$\leftrightarrow$Japanese translation datasets.

 For Chinese$\leftrightarrow$English translation, our training data consists of 2.08M sentence pairs extracted from LDC corpus{\footnote[1]{The corpora include LDC2000T50, LDC2002T01, LDC2002E18, LDC2003E07, LDC2003E14, LDC2003T17 and LDC2004T07.}}. We use NIST 2003 (MT03) dataset as the validation set, NIST 2004 (MT 04), NIST 2005 (MT05), NIST 2006 (MT 06) datasets as our test sets. We use BPE ~\cite{Sennrich:2016A} to encode Chinese and English respectively, and limit the source and target vocabularies to the most frequent 30K tokens.

%

For English$\leftrightarrow$Japanese translation tasks, we use KFTT datasets{\footnote[2]{http://isw3.naist.jp/~philip-a/emnlp2016/}} consisting of 440K sentence pairs, which also is used in \cite{arthur2016incorporating}. Sentences were encoded using BPE, and the vocabulary sizes are 31K and 33K for English and Japanese respectively.

\subsection{Training Details}

We use the tensor2tensor{\footnote[3]{https://github.com/tensorflow/tensor2tensor}} library for training and evaluating our Transformer model. Additionally, we utilize the OpenNMT{\footnote[4]{https://github.com/OpenNMT/OpenNMT-py}} to train and test our RNMT model.

For our Transformer model, we employ the Adam optimizer with $\beta_1$=0.9, $\beta_2$=0.98, and $\epsilon$=$10^{-9}$. We use the same warmup and decay strategy for learning rate as \cite{vaswani2017attention}, with 4,000 warmup steps. During training, we employ label smoothing of value $\epsilon_{ls}$=0.1. For evaluation, we use beam search with a beam size of $k$=4 and length penalty $\alpha$=0.6. Additionally, we use 6 encoder and decoder layers, hidden state size $d_x$=1024, 16 attention-heads, and 4096 feed forward inner-layer dimensions.

For RNMT, we use 4 encoder and decoder layers with LSTM. The word embedding dimension and the size of hidden layers are both set to 1,000. we use global attention ~\cite{Luong:2015A} and beam search with beam size $k$=12 in RNMT. Parameter optimization is performed using Adam with the default configuration.

\section{Results and Analysis}

Below we discuss the results of our translation experiments about representor and multi-task learning framework, measuring translation quality with case-insensitive BLEU~\cite{Papineni:2002}.



\subsection{Language-Independent Representor}

We first analyze the effects of representor on both RNMT and Transformer.
The results on Chinese-English translation are shown in Table~\ref{resultWS}.
Experimental results demonstrate that our proposed \emph{frequency-based embedding weight sharing} strategy  obtains significant accuracy improvements on RNMT (39.73 vs. 38.92) and Transformer (47.19 vs. 46.74).
We find that the weight sharing of encoder and decoder is more effective to Transformer (46.39 vs. 46.74) than RNMT (37.19 vs 38.92), where the number of Transformer parameters is sharply reduced, and their performance is comparable to baseline.
Note that even though our proposed model (Transformer + ES + EDS) only uses 39.5\% parameters of baseline model, their performance is comparable.

Furthermore, we also compare with the base Transformer model{\footnote[5]{For base model, we use hidden state size $d_x$=512, 8 attention-heads, and 2048 feed forward inner-layer dimensions.}} and two layers encoder-decoder Transformer model, whose results are reported in the last two lines of Table~\ref{resultWS}.
The base Transformer reduces model parameters but decreases dramatically the translation quality.
Our proposed weight sharing models outperform the two layer Transformer in terms of BLEU scores and model parameters.
Considering the balance between model size and translation performance, we will use this kind of representor without LS (Transformer + ES + EDS) in subsequent experiments.

\begin{table}
\setlength{\belowcaptionskip}{-0.5cm}
\centering
\begin{tabular}{l|c|c}
  \hline
   Model                                           &    Params             &      Test         \\
  \hline
  \hline
  \citeauthor{arthur2016incorporating}~\shortcite{arthur2016incorporating}     &   -                      &  23.20   \\
  \citeauthor{N18-1031}~\shortcite{N18-1031}     &   -                      &  26.20   \\
  \hline
  Transformer                                            &  210.8M                 &  29.88 \\
  Our Representor                                       &  110.0M               &  \textbf{31.63}            \\
  \hline
\end{tabular}
\caption{Translation performance of Transformer and our proposed representor on English-Japanese.
    We also provide some experimental results of the first two models on the same data set.
  } \label{resultWS2}
\end{table}

Table~\ref{resultWS2} shows the performance of representor on low-resource English-Japanese translation, which demonstrates that even though the weight sharing models contain fewer parameters than the baseline models, it gets an improvement of +1.75 BLEU points.
Furthermore, we show that our representor can reduce the size of Transformer models to less than half of their original size while achieving significant improvement for low-resource translation.

\begin{table*}
	\setlength{\belowcaptionskip}{-0.5cm}
	\newcommand{\tabincell}[2]{\begin{tabular}{@{}#1@{}}#2\end{tabular}}
	\centering
	\begin{tabular}{p{2cm}|c||cccc}
		\hline
		\multirow{2}{*}{System}&\multirow{2}{*}{Decoding Manner}&   \multicolumn{2}{c}{Chinese$\leftrightarrow$English}       &   \multicolumn{2}{c}{English$\leftrightarrow$Japanese}   \\
		\cmidrule(lr){3-4}  \cmidrule(lr){5-6}
		&                       &   Ch$\rightarrow$ En& En$\rightarrow$ Ch        &   En$\rightarrow$ Ja     &   Ja$\rightarrow$ En     \\
		\hline
		\hline
		Transformer    &       Left-to-right   &     46.74       &    22.49      &   29.88    &   23.70       \\  
		Representor        &       Left-to-right     &        46.71     &     21.74             &   31.63    &  25.61    \\  
		S-T\&T-S   &       Left-to-right   &     46.72      &    21.12             &   32.81$\dagger$         &      25.88$\dagger$     \\  
		L-R\&R-L  &   Mixed Decoding   &   47.41$\ddagger$  &  \textbf{23.51}$\dagger$   &   32.84$\dagger$    &  25.55$\dagger$  \\ 
		\multirow{2}{*}{CFP Method}   & Mixed Decoding  & 47.25$\ddagger$   &    23.08$\ddagger$   &  32.62$\dagger$   &   \textbf{26.59}$\dagger$  \\
		& Joint Decoding  &  \textbf{48.08}$\dagger$  & 23.47$\dagger$    &  \textbf{33.41}$\dagger$  &  26.19$\dagger$   \\
		\hline
	\end{tabular}
	\caption{Experimental results of multi-task learning framework for four directions on two translation tasks.
		S-T\&T-S  and L-R\&R-L extend represnetor by introducing the new training objective Eq.(\ref{t2s-patterns}) and Eq.(\ref{r2l-patterns}) respectively.
		The CFP method means the multi-task learning technique that utilizes a representor to training four translation directions shown in Eq.(\ref{four-patterns}).
		The results marked by $\dagger$ are significantly better than Transformer (p $<$ 0.01), and ``$\ddagger$" denotes p $<$ 0.05.
	} \label{datatrans}
\end{table*}

\subsection{Multi-Task Learning for Representor}

In this section, we will report and analyze the results of multi-task learning framework, which conducts multiple translation tasks in a language-independent representor.


\textbf{Results on Chinese$\leftrightarrow$English}
Our results of multi-task learning technique on  large-scale Chinese$\leftrightarrow$English translation tasks are presented in Table~\ref{datatrans}.
We find that our CFP model with joint decoding obtains the best results in Ch-En translation, and it outperforms standard Transformer by 1.34 BLEU points.
Additionally, L-R\&R-L model, getting an improvement of 1.02 BLEU points than Transformer, behaves better than CFP model on En-Ch translation.
Experiments demonstrate that our single model can conduct source-to-target and target-to-source translation by using less than a quarter of parameters, while still achieving better performance on large-scale datasets.

\textbf{Results on English$\leftrightarrow$Japanese}
Three kinds of task-level weight sharing methods have achieved remarkable improvements in low-resource translation, as demonstrated in Table~\ref{datatrans}.
Experiments show that our CFP framework combining four translation patterns outperforms the Transformer by +3.53 and + 2.89 BLEU points in bidirectional English-Japanese translation separately.
We think the main reason is that our proposed framework can be regarded as a data augmentation technique for low-resource translation.


\begin{figure}[h]
\setlength{\abovecaptionskip}{-0.01cm}
\setlength{\belowcaptionskip}{-0.3cm}
    \centering
    \includegraphics[width=7cm]{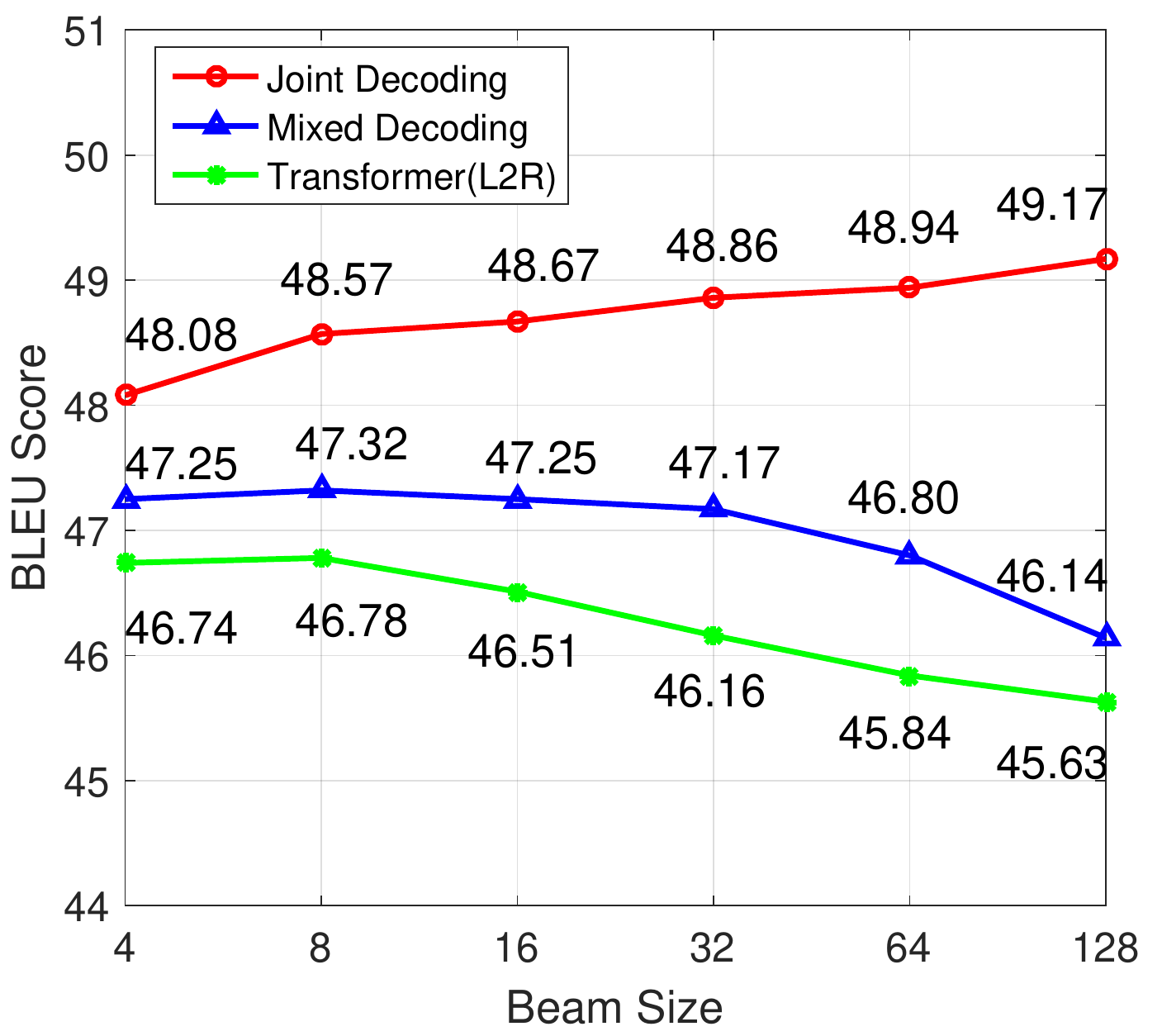}
    \caption{Translation qualities (BLEU score) of our joint decoding, mixed decoding, and Transformer(L2R) on Chinese-English translation as beam size become larger.
    } \label{beam}
\end{figure}

\textbf{Effect of Large Decoding Space}
\cite{tu2017neural} observed that general likelihood objective favors short translations, and can not decode well in a large search space.
We present the effect of CFP model with mixed and joint decoding on different beam sizes k, as shown in Figure \ref{beam}. Unlike Transformer, increasing the size of decoding space leads to improving the BLEU scores for our joint decoding.
Although joint decoding is more complicated than mixed decoding, it can capture dependency of four translation directions to select the best candidate.


%

\textbf{Ratio of L2R and R2L}
The mixed decoding allows the model to choose either right-to-left or left-to-right decoding based on the particular source representation. We have conducted the statistical analysis on the proportion of left-to-right decoding and right-to-left decoding. We find that the proportion of L2R in Ch-En translation is higher than that of R2L, and En-Ja has opposite phenomenon. These phenomena are consistent with the experimental results that L2R decoding behaves better than R2L decoding in Ch-En and L2R decoding is worse than R2L decoding in En-Ja. 


\begin{table}[h]
\setlength{\belowcaptionskip}{-0.3cm}
\centering
\begin{tabular}{l|cc|cc}
  \hline
  \multirow{2}{*}{Direction}  &\multicolumn{2}{c}{Chinese$\leftrightarrow$English}        &\multicolumn{2}{c}{English$\leftrightarrow$Japanese}    \\  \cmidrule(lr){2-3} \cmidrule(lr){4-5}
                          &    Ch$\rightarrow$En        & En$\rightarrow$Ch    &   En$\rightarrow$Ja   &  Ja$\rightarrow$En    \\
  \hline
  L2R    &  56.0\%        &   54.8\%    &    48.2\%   &    45.3\%     \\
  R2L    &  44.0\%      &   45.2\%      &     51.8\%   &     54.7\%    \\
  \hline
\end{tabular}
\caption{Translation proportion of L2R and R2L manners on Ch$\leftrightarrow$En and En$\leftrightarrow$Ja translation tasks.
  } \label{l2r&r2l}
\end{table}

\begin{table*}
\setlength{\belowcaptionskip}{-0.5cm}
\centering
\begin{tabular}{p{2.6cm}p{12.1cm}}
  \hline
  Source  & \heng2\fu2 \xia4\fang1 \shi4 \yi1 \zhang1 \shi2\duo1 \ping2\fang1\mi3 de \wang2\xuan3 \ju4\fu2 \yi2\xiang4, \ta1 \dai4 zhe \yan3\jing4, \mian4\lu4 \wei1\xiao4, \qi4\zhi4 \ru2\ya3。\\
  Reference &  \dashuline{below the banner , there was a big picture of \textbf{wang xuan} that was more than ten square meters in size}  . \textcolor{red}{ \textit{wang was wearing a pair of glasses , smiling with a scholarly air .} } \\
  \hline
  \hline
  Transformer(L2R) &  \dashuline{below the scroll is a huge portrait of a 10-square-meter wang} .   \\
  Transformer(R2L) &   \textcolor{red}{ \emph{he had a look on his face , wearing glasses , a smiling face and a refined air} } .\\
  Mixed Decoding &   \dashuline{below the banner is a big portrait of more than 10 square meters}  , \textcolor{red}{ \emph{wearing glasses , smiling , and elegant .} }   \\
  Joint Decoding &  \dashuline{below the banner is a huge portrait of \textbf{wang xuan} , which is more than 10 square meters long }.   \textcolor{red}{ \emph{he wears glasses , smiles , and is elegant .} }  \\
  \hline
\end{tabular}
  \caption{Chinese-English translation examples of Transformer decoding in left-to-right and right-to-left ways, our proposed models using mixed decoding (MD) and joint decoding (JD) technique respectively.} \label{example}
\end{table*}

\textbf{Length Analysis}
We follow ~\citeauthor{Bahdanau:2015}~\shortcite{Bahdanau:2015} to group sentences of similar lengths together and compute a BLEU score per
group.
Figure ~\ref{length} shows that joint decoding can alleviate the inability of translating long sentences, but mixed decoding behaves better in handling short sentences.

\begin{figure}[h]
\setlength{\abovecaptionskip}{-0.01cm}
\setlength{\belowcaptionskip}{-0.3cm}
    \centering
    \includegraphics[width=7.2cm]{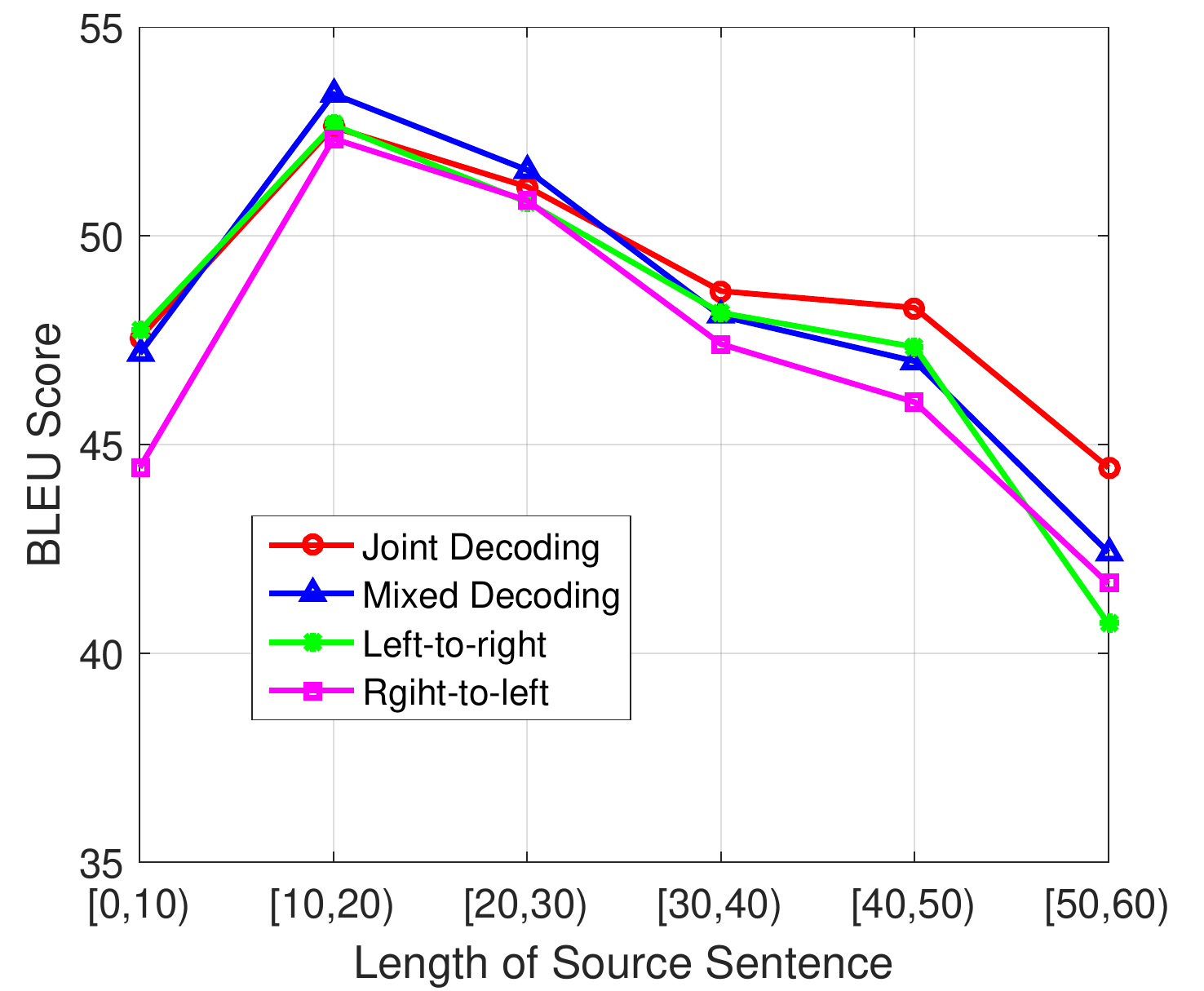}
    \caption{Length Analysis - performance of translations with respect to the lengths of the source sentences.
    } \label{length}
\end{figure}

\textbf{Duality Analysis}
Table~\ref{example} gives a translation example of different models.
Transformer(L2R) drops out the second half of the Chinese sentence, while Transformer(R2L) omits the versa part.
Mixed decoding gets better translation result than Transformer.
Compared with mixed decoding, joint decoding not only translates successfully the latter part of the sentence, but also translates accurately the name \wang2\xuan3, which demonstrates that our proposed models can enhance the  translation duality and agreement.

\section{Related Work}

Our work is inspired by three lines of research on improving NMT by:

\textbf{Model Compression and Multi-Task Learning }
To reduce model parameters, weight pruning and knowledge distillation have been proposed to compress NMT models~\cite{see2016compression,kim2016sequence}.
Additionally, a recent lines of research have investigated multilingual machine translation by using multi-task learning~\cite{ha2016toward,firat2016zero,johnson2016google}.
However, 
we achieve model compression from a very different perspective, even though this is not the main purpose. Besides, our model can also be adapted to these approaches because they are orthogonal to each other.

\textbf{Data Augmentation for NMT}
For low-resource NMT, most of the existing approaches and models mainly focus on utilizing transfer learning ~\cite{zoph2016transfer,lakew2017multilingual,gu2018universal}, or exploiting large-scale monolingual data ~\cite{Cheng:2016,Sennrich:2016B,Zhang:2016B}.
\citeauthor{fadaee2017data}~\shortcite{fadaee2017data} proposed a data augmentation method that generates new sentence pairs by replacing high-frequency words with rare words.
In this paper, our proposed framework can be viewed as a novel data augmentation technique that expands  training corpora multiple times by data transformation for low-resource translation.

\textbf{Translation Duality for NMT }
Some work have noticed the symmetry of translation ~\cite{Cheng:2016,he2016dual,zhang2018joint}, which attempt to bridge source-to-target translation and target-to-source translation.
In the other hand, many studies have pointed out the shortcoming of unidirectional decoding, and proposed some approaches to combine the advantage of left-to-right decoding and right-to-left decoding ~\cite{liu2016agreementa,sennrich2017university,hassan2018achieving}.
However, above methods are designed for alternative translation agreement and use two different encoder-decoder models.
We attempt at designing a unified framework to boost the duality of four translation directions by using one representor.

\section{Conclusions and Future Work }

In this work, we propose a noval language-independent NMT framework in which a language-independent representor can perform multiple translation tasks by using weight sharing and multi-task learning.
Our proposed framework can drastically reduce model parameters and  take full advantage of language duality.
Experiments on two resource-rich and low-resource translation tasks show that our framework can  use only a quarter of parameters while achieving significant improvements over conventional NMT models.
For future work, we plan to design explicit training constraints in the multi-task learning framework to further exploit the language duality.
Additionally, it is interesting to extend this approach to monolingual data utilization and unsupervised neural machine translation.


\bibliography{aaai2019}

\begin{thebibliography}{}

\bibitem[\protect\citeauthoryear{Arthur, Neubig, and
  Nakamura}{2016}]{arthur2016incorporating}
Arthur, P.; Neubig, G.; and Nakamura, S.
\newblock 2016.
\newblock Incorporating discrete translation lexicons into neural machine
  translation.
\newblock In {\em Proceedings of the 2016 Conference on Empirical Methods in
  Natural Language Processing},  1557--1567.
\newblock Association for Computational Linguistics.

\bibitem[\protect\citeauthoryear{Bahdanau, Cho, and
  Bengio}{2015}]{Bahdanau:2015}
Bahdanau, D.; Cho, K.; and Bengio, Y.
\newblock 2015.
\newblock {\em Neural machine translation by jointly learning to align and
  translate}.
\newblock In Proceedings of ICLR 2015.

\bibitem[\protect\citeauthoryear{Bojar \bgroup et al\mbox.\egroup
  }{2017}]{bojar2017findings}
Bojar, O.; Chatterjee, R.; Federmann, C.; Graham, Y.; Haddow, B.; Huang, S.;
  Huck, M.; Koehn, P.; Liu, Q.; Logacheva, V.; et~al.
\newblock 2017.
\newblock Findings of the 2017 conference on machine translation (wmt17).
\newblock In {\em Proceedings of the Second Conference on Machine Translation},
   169--214.

\bibitem[\protect\citeauthoryear{Cheng \bgroup et al\mbox.\egroup
  }{2016a}]{cheng2015agreement}
Cheng, Y.; Shen, S.; He, Z.; He, W.; Wu, H.; Sun, M.; and Liu, Y.
\newblock 2016a.
\newblock Agreement-based joint training for bidirectional attention-based
  neural machine translation.
\newblock In {\em In Proceedings of IJCAI 2016}.

\bibitem[\protect\citeauthoryear{Cheng \bgroup et al\mbox.\egroup
  }{2016b}]{Cheng:2016}
Cheng, Y.; Xu, W.; He, Z.; He, W.; Wu, H.; Sun, M.; and Liu, Y.
\newblock 2016b.
\newblock {\em Semi-supervised learning for neural machine translation}.
\newblock In Proceedings of ACL 2016.

\bibitem[\protect\citeauthoryear{Cho \bgroup et al\mbox.\egroup
  }{2014}]{Cho:2014}
Cho, K.; van Merrienboer, B.; Gulcehre, C.; Bahdanau, D.; Bougares, F.;
  Schwenk, H.; and Bengio, Y.
\newblock 2014.
\newblock {\em Learning phrase representations using RNN encoder–decoder for
  statistical machine translation}.
\newblock In Proceedings of EMNLP 2014.

\bibitem[\protect\citeauthoryear{Fadaee, Bisazza, and
  Monz}{2017}]{fadaee2017data}
Fadaee, M.; Bisazza, A.; and Monz, C.
\newblock 2017.
\newblock Data augmentation for low-resource neural machine translation.
\newblock In {\em Proceedings of the 55th Annual Meeting of the Association for
  Computational Linguistics (Volume 2: Short Papers)},  567--573.
\newblock Association for Computational Linguistics.

\bibitem[\protect\citeauthoryear{Firat \bgroup et al\mbox.\egroup
  }{2016}]{firat2016zero}
Firat, O.; Sankaran, B.; Al-Onaizan, Y.; Vural, F. T.~Y.; and Cho, K.
\newblock 2016.
\newblock {\em Zero-resource translation with multi-lingual neural machine
  translation}.
\newblock In Proceedings of EMNLP 2016.

\bibitem[\protect\citeauthoryear{Gehring \bgroup et al\mbox.\egroup
  }{2017}]{gehring2017convolutional}
Gehring, J.; Auli, M.; Grangier, D.; Yarats, D.; and Dauphin, Y.
\newblock 2017.
\newblock Convolutional sequence to sequence learning.
\newblock In {\em ICML}.

\bibitem[\protect\citeauthoryear{Gu \bgroup et al\mbox.\egroup
  }{2018}]{gu2018universal}
Gu, J.; Hassan, H.; Devlin, J.; and Li, V.~O.
\newblock 2018.
\newblock Universal neural machine translation for extremely low resource
  languages.
\newblock {\em In Proceedings of NAACL 2018}.

\bibitem[\protect\citeauthoryear{Ha, Niehues, and Waibel}{2016}]{ha2016toward}
Ha, T.-L.; Niehues, J.; and Waibel, A.
\newblock 2016.
\newblock Toward multilingual neural machine translation with universal encoder
  and decoder.
\newblock {\em In Proceedings of IWSLT 2016}.

\bibitem[\protect\citeauthoryear{Hassan \bgroup et al\mbox.\egroup
  }{2018}]{hassan2018achieving}
Hassan, H.; Aue, A.; Chen, C.; Chowdhary, V.; Clark, J.; Federmann, C.; Huang,
  X.; Junczys-Dowmunt, M.; Lewis, W.; Li, M.; et~al.
\newblock 2018.
\newblock Achieving human parity on automatic chinese to english news
  translation.
\newblock {\em arXiv preprint arXiv:1803.05567}.

\bibitem[\protect\citeauthoryear{He \bgroup et al\mbox.\egroup
  }{2016}]{he2016dual}
He, D.; Xia, Y.; Qin, T.; Wang, L.; Yu, N.; Liu, T.; and Ma, W.-Y.
\newblock 2016.
\newblock Dual learning for machine translation.
\newblock In {\em Advances in Neural Information Processing Systems},
  820--828.

\bibitem[\protect\citeauthoryear{Hochreiter and
  Schmidhuber}{1997}]{hochreiter1997long}
Hochreiter, S., and Schmidhuber, J.
\newblock 1997.
\newblock Long short-term memory.
\newblock {\em Neural computation} 9(8):1735--1780.

\bibitem[\protect\citeauthoryear{Johnson \bgroup et al\mbox.\egroup
  }{2017}]{johnson2016google}
Johnson, M.; Schuster, M.; Le, Q.~V.; Krikun, M.; Wu, Y.; Chen, Z.; Thorat, N.;
  Vi{\'e}gas, F.; Wattenberg, M.; Corrado, G.; et~al.
\newblock 2017.
\newblock Google's multilingual neural machine translation system: enabling
  zero-shot translation.
\newblock {\em Transactions of the Association for Computational Linguistics}.

\bibitem[\protect\citeauthoryear{Kim and Rush}{2016}]{kim2016sequence}
Kim, Y., and Rush, A.~M.
\newblock 2016.
\newblock Sequence-level knowledge distillation.
\newblock In {\em Proceedings of the 2016 Conference on Empirical Methods in
  Natural Language Processing},  1317--1327.
\newblock Association for Computational Linguistics.

\bibitem[\protect\citeauthoryear{Koehn and Knowles}{2017}]{koehn2017six}
Koehn, P., and Knowles, R.
\newblock 2017.
\newblock Six challenges for neural machine translation.
\newblock In {\em Proceedings of the First Workshop on Neural Machine
  Translation},  28--39.
\newblock Association for Computational Linguistics.

\bibitem[\protect\citeauthoryear{Lakew, Gangi, and
  Federico}{2017}]{lakew2017multilingual}
Lakew, S.~M.; Gangi, M. A.~D.; and Federico, M.
\newblock 2017.
\newblock Multilingual neural machine translation for low resource languages.
\newblock In {\em CLiC-it}.

\bibitem[\protect\citeauthoryear{Liu \bgroup et al\mbox.\egroup
  }{2016}]{liu2016agreementa}
Liu, L.; Utiyama, M.; Finch, A.; and Sumita, E.
\newblock 2016.
\newblock Agreement on target-bidirectional neural machine translation.
\newblock In {\em Proceedings of the 2016 Conference of the North American
  Chapter of the Association for Computational Linguistics: Human Language
  Technologies},  411--416.

\bibitem[\protect\citeauthoryear{Luong, Pham, and Manning}{2015}]{Luong:2015A}
Luong, M.-T.; Pham, H.; and Manning, C.~D.
\newblock 2015.
\newblock {\em Effective approaches to attention-based neural machine
  translation}.
\newblock In Proceedings of EMNLP 2015.

\bibitem[\protect\citeauthoryear{Nguyen and Chiang}{2018}]{N18-1031}
Nguyen, T., and Chiang, D.
\newblock 2018.
\newblock Improving lexical choice in neural machine translation.
\newblock In {\em Proceedings of the 2018 Conference of the North American
  Chapter of the Association for Computational Linguistics: Human Language
  Technologies, Volume 1 (Long Papers)},  334--343.
\newblock Association for Computational Linguistics.

\bibitem[\protect\citeauthoryear{Papineni \bgroup et al\mbox.\egroup
  }{2002}]{Papineni:2002}
Papineni, K.; Roukos, S.; Ward, T.; and Zhu, W.
\newblock 2002.
\newblock {\em Bleu: a methof for automatic evaluation of machine translation}.
\newblock In Proceedings of ACL 2002.

\bibitem[\protect\citeauthoryear{Press and Wolf}{2017}]{press2016using}
Press, O., and Wolf, L.
\newblock 2017.
\newblock Using the output embedding to improve language models.
\newblock In {\em Proceedings of the 15th Conference of the European Chapter of
  the Association for Computational Linguistics: Volume 2, Short Papers},
  157--163.
\newblock Association for Computational Linguistics.

\bibitem[\protect\citeauthoryear{See, Luong, and
  Manning}{2016}]{see2016compression}
See, A.; Luong, M.-T.; and Manning, C.~D.
\newblock 2016.
\newblock Compression of neural machine translation models via pruning.
\newblock In {\em Proceedings of The 20th SIGNLL Conference on Computational
  Natural Language Learning},  291--301.
\newblock Association for Computational Linguistics.

\bibitem[\protect\citeauthoryear{Sennrich \bgroup et al\mbox.\egroup
  }{2017}]{sennrich2017university}
Sennrich, R.; Birch, A.; Currey, A.; Germann, U.; Haddow, B.; Heafield, K.;
  Barone, A. V.~M.; and Williams, P.
\newblock 2017.
\newblock The university of edinburgh's neural mt systems for wmt17.
\newblock {\em arXiv preprint arXiv:1708.00726}.

\bibitem[\protect\citeauthoryear{Sennrich, Haddow, and
  Birch}{2016a}]{Sennrich:2016B}
Sennrich, R.; Haddow, B.; and Birch, A.
\newblock 2016a.
\newblock {\em Improving neural machine translation models with monolingual
  data}.
\newblock In Proceedings of ACL 2016.

\bibitem[\protect\citeauthoryear{Sennrich, Haddow, and
  Birch}{2016b}]{Sennrich:2016A}
Sennrich, R.; Haddow, B.; and Birch, A.
\newblock 2016b.
\newblock {\em Neural machine translation of rare words with subword units}.
\newblock In Proceedings of ACL 2016.

\bibitem[\protect\citeauthoryear{Tu \bgroup et al\mbox.\egroup
  }{2017}]{tu2017neural}
Tu, Z.; Liu, Y.; Shang, L.; Liu, X.; and Li, H.
\newblock 2017.
\newblock Neural machine translation with reconstruction.
\newblock In {\em AAAI},  3097--3103.

\bibitem[\protect\citeauthoryear{Vaswani \bgroup et al\mbox.\egroup
  }{2017}]{vaswani2017attention}
Vaswani, A.; Shazeer, N.; Parmar, N.; Uszkoreit, J.; Jones, L.; Gomez, A.~N.;
  Kaiser, {\L}.; and Polosukhin, I.
\newblock 2017.
\newblock Attention is all you need.
\newblock In {\em Advances in Neural Information Processing Systems},
  6000--6010.

\bibitem[\protect\citeauthoryear{Wu \bgroup et al\mbox.\egroup
  }{2016}]{Wu:2016}
Wu, Y.; Schuster, M.; Chen, Z.; Le, Q.~V.; and Mohammad~Norouzi, e.~a.
\newblock 2016.
\newblock {\em Google’s neural machine translation system: bridging the gap
  between human and machine translation}.
\newblock arXiv preprint arXix:1609.08144.

\bibitem[\protect\citeauthoryear{Zhang and Zong}{2016}]{Zhang:2016B}
Zhang, J., and Zong, C.
\newblock 2016.
\newblock {\em Exploiting source-side monolingual data in neural machine
  translation.}
\newblock In Proceedings of EMNLP 2016.

\bibitem[\protect\citeauthoryear{Zhang \bgroup et al\mbox.\egroup
  }{2018}]{zhang2018joint}
Zhang, Z.; Liu, S.; Li, M.; Zhou, M.; and Chen, E.
\newblock 2018.
\newblock Joint training for neural machine translation models with monolingual
  data.
\newblock {\em In Proceedings of AAAI 2018}.

\bibitem[\protect\citeauthoryear{Zhou \bgroup et al\mbox.\egroup
  }{2016}]{zhou2016deep}
Zhou, J.; Cao, Y.; Wang, X.; Li, P.; and Xu, W.
\newblock 2016.
\newblock Deep recurrent models with fast-forward connections for neural
  machine translation.
\newblock {\em Transactions of the Association for Computational Linguistics}
  4:371--383.

\bibitem[\protect\citeauthoryear{Zoph \bgroup et al\mbox.\egroup
  }{2016}]{zoph2016transfer}
Zoph, B.; Yuret, D.; May, J.; and Knight, K.
\newblock 2016.
\newblock Transfer learning for low-resource neural machine translation.
\newblock In {\em Proceedings of the 2016 Conference on Empirical Methods in
  Natural Language Processing},  1568--1575.
\newblock Association for Computational Linguistics.

\end{thebibliography}
\bibliographystyle{aaai}

\end{document}